%% file: main.tex
\documentclass[runningheads]{llncs}

\usepackage[mobile]{eccv}

\usepackage{eccvabbrv}

\usepackage{graphicx}
\usepackage{booktabs}

\usepackage[accsupp]{axessibility}  

\usepackage[pagebackref,breaklinks,colorlinks,citecolor=eccvblue]{hyperref}

\usepackage{orcidlink}

\input{lib/my_instructions.tex}

\begin{document}

\title{LocalMamba: Visual State Space Model with Windowed Selective Scan} 

\titlerunning{LocalMamba}

\author{Tao Huang\inst{1} \and Xiaohuan Pei\inst{1} \and \\ Shan You\inst{2} \and Fei Wang\inst{3} \and Chen Qian\inst{2} \and Chang Xu\inst{1}}

\authorrunning{T. Huang et al.}

\institute{School of Computer Science, Faculty of Engineering, The University of Sydney \and
SenseTime Research \and University of Science and Technology of China
}

{\def\thefootnote{}\footnotetext{Correspondence to: Tao Huang <thua7590@uni.sydney.edu.au>, Shan You <youshan@sensetime.com>, Chang Xu <c.xu@sydney.edu.au>}}

\maketitle

\begin{abstract}
  Recent advancements in state space models, notably Mamba, have demonstrated significant progress in modeling long sequences for tasks like language understanding. Yet, their application in vision tasks has not markedly surpassed the performance of traditional Convolutional Neural Networks (CNNs) and Vision Transformers (ViTs). This paper posits that the key to enhancing Vision Mamba (ViM) lies in optimizing scan directions for sequence modeling. Traditional ViM approaches, which flatten spatial tokens, overlook the preservation of local 2D dependencies, thereby elongating the distance between adjacent tokens. We introduce a novel local scanning strategy that divides images into distinct windows, effectively capturing local dependencies while maintaining a global perspective. Additionally, acknowledging the varying preferences for scan patterns across different network layers, we propose a dynamic method to independently search for the optimal scan choices for each layer, substantially improving performance. Extensive experiments across both plain and hierarchical models underscore our approach's superiority in effectively capturing image representations. For example, our model significantly outperforms Vim-Ti by 3.1\% on ImageNet with the same 1.5G FLOPs. Code is available at: \url{https://github.com/hunto/LocalMamba}.
\end{abstract}

\section{Introduction}
\label{sec:intro}

Structured State Space Models (SSMs) have recently gained prominence as a versatile architecture in sequence modeling, heralding a new era of balancing computational efficiency and model versatility \cite{gu2021combining,gu2022efficiently,mehta2023long,gu2023mamba}. These models synthesize the best attributes of Recurrent Neural Networks (RNNs) and Convolutional Neural Networks (CNNs), drawing inspiration from the foundational principles of classical state space models \cite{kalman1960new}. Characterized by their computational efficiency, SSMs exhibit linear or near-linear scaling complexity with sequence length, making them particularly suited for handling long sequences. Following the success of Mamba \cite{gu2023mamba}, a novel variant that incorporates selective scanning (S6), there has been a surge in applying SSMs to a wide range of vision tasks. These applications extend from developing generic foundation models \cite{zhu2024vision,liu2024vmamba} to advancing fields in image segmentation \cite{ruan2024vm,liu2024swin,xing2024segmamba,ma2024u} and synthesis \cite{guo2024mambair}, demonstrating the model's adaptability and potential in visual domain.

\begin{figure}[t]
    \centering
    \includegraphics[width=\linewidth]{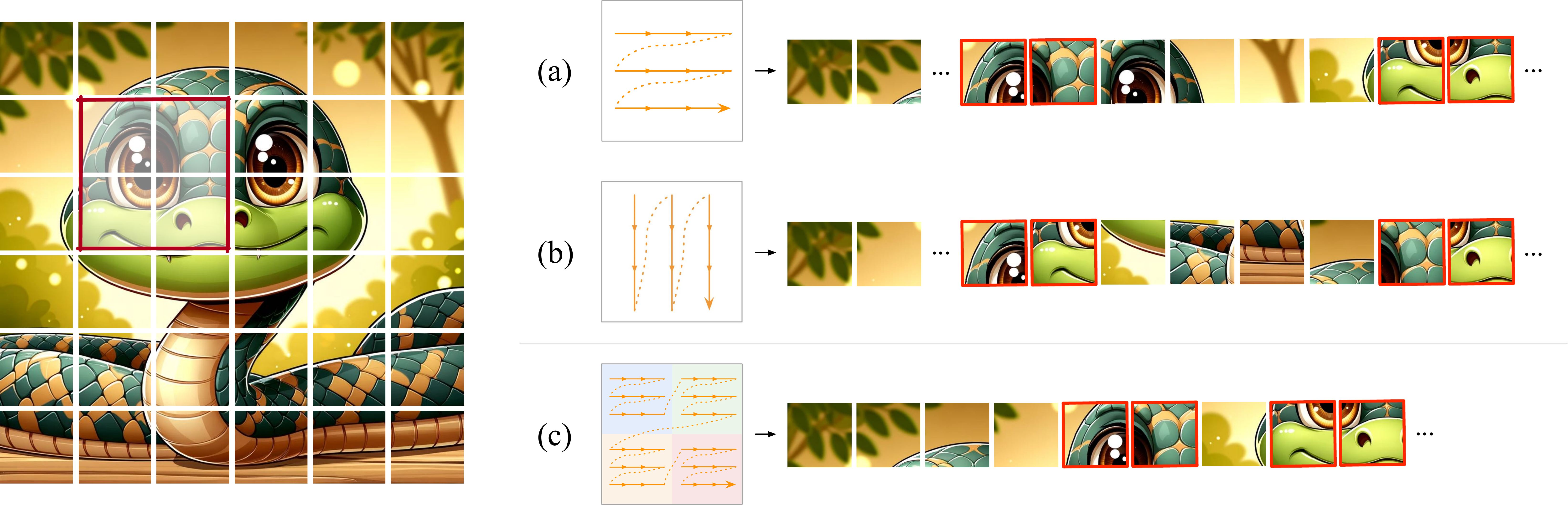}
    \caption{Illustration of scan methods. (a) and (b): Previous methods Vim \cite{zhu2024vision} and VMamba \cite{liu2024vmamba} traverse the entire row or column axis, resulting in significant distances for capturing dependencies between neighboring pixels within the same semantic region (\eg, the left eye in the image). (c) We introduce a novel scan method that partitions tokens into distinct windows, facilitating traversal within each window (window size is $3\times 3$ here). This approach enhances the ability to capture local dependencies.}
    \label{fig:scan_illustration}
    \vspace{-4mm}
\end{figure}

\begin{wrapfigure}{r}{0.35\linewidth}
    \vspace{-8mm}
    \centering
    \includegraphics[width=0.8\linewidth,height=0.85\linewidth]{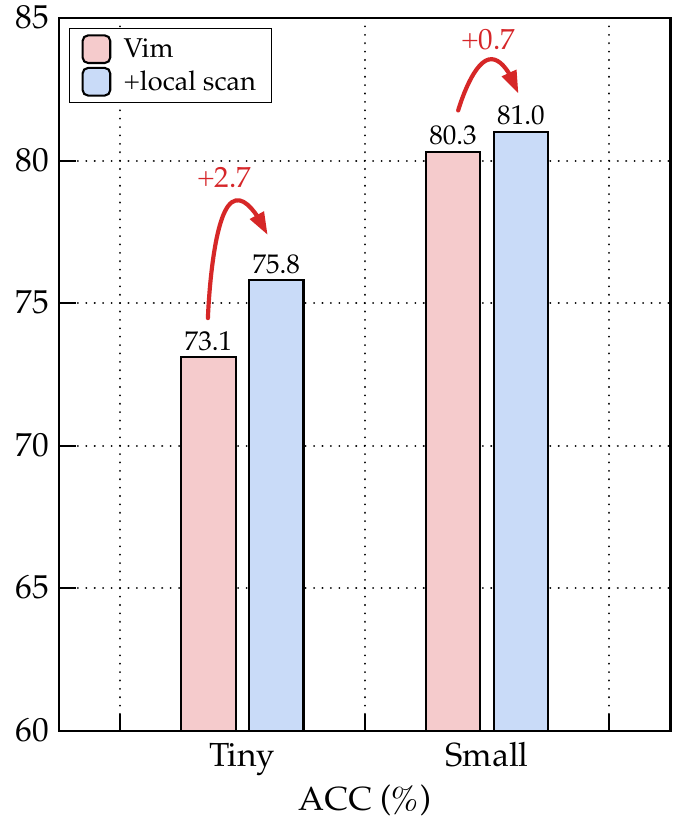}
    \vspace{-2mm}
    \caption{By extending the original scan with our local scan mechanism, our method significantly improves the ImageNet accuracies of Vim \cite{zhu2024vision} while keeping similar FLOPs.}
    \label{fig:cmp_vim}
    \vspace{-8mm}
\end{wrapfigure}
Typically these vision studies need to transform 2D images into 1D sequences for SSM-based processing, and then integrate the original SSM structure of Mamba into their foundational models for specific tasks. Nevertheless, they have only shown modest improvements over traditional CNNs \cite{krizhevsky2012imagenet,simonyan2014very,he2016deep,xie2017aggregated,sandler2018mobilenetv2,radosavovic2020designing} and Vision Transformers (ViTs) \cite{dosovitskiy2021an,liu2021swin,chu2021twins,touvron2021training}. This modest advancement underscores a significant challenge: the non-causal nature of 2D spatial pattern in images is inherently at odds with the causal processing framework of SSMs. As illustrated in Figure \ref{fig:scan_illustration}, traditional methods that flatten spatial data into 1D tokens disrupt the natural local 2D dependencies, weakening the model's ability to accurately interpret spatial relationships. Although VMamba \cite{liu2024vmamba} introduces a 2D scanning technique to address this by scanning images in both horizontal and vertical directions, it still struggles with maintaining the proximity of originally adjacent tokens within the scanned sequences, which is critical for effective local representation modeling.

In this work, we introduce a novel approach to improve local representation within Vision Mamba (ViM) by segmenting the image into multiple distinct local windows. Each window is scanned individually before conducting a traversal across windows, ensuring that tokens within the same 2D semantic region are processed closely together. This method significantly boosts the model's capability to capture details among local regions, with the experimental results validated in Figure \ref{fig:cmp_vim}. We design our foundational block by integrating both traditional global scanning directions and our novel local scanning technique, empowering the model to assimilate comprehensive global and nuanced local information. Furthermore, to better aggregate features from these diverse scanning processes, we propose a spatial and channel attention module, SCAttn, engineered to discern and emphasize valuable information while filtering out redundancy.

Acknowledging the distinct impact of scanning directions on feature representation (for instance, a local scan with a window size of $3$ excels in capturing smaller objects or details, whereas a window size of $7$ is better suited for larger objects), we introduce a direction search method for selecting optimal scanning directions. This variability is especially pronounced across different layers and network depths. Inspired by DARTS \cite{liu2018darts}, we proceed from discrete selection to a continuous domain, represented by a learnable factor, to incorporate multiple scanning directions within a single network. After the training of this network, the most effective scanning directions are determined by identifying those with the highest assigned probabilities.

Our developed models, LocalVim and LocalVMamba, incorporate both plain and hierarchical structures, resulting in notable enhancements over prior methods. Key contributions of this study include:
\begin{enumerate}
    \item We introduce a novel scanning methodology for SSMs that includes localized scanning within distinct windows, significantly enhancing our models' ability to capture detailed local information in conjunction with global context.
    \item We develop a method for searching scanning directions across different network layers, enabling us to identify and apply the most effective scanning combinations, thus improving network performance.
    \item We present two model variants, designed with plain and hierarchical structures. Through extensive experimentation on image classification, object detection, and semantic segmentation tasks, we demonstrate that our models achieve significant improvements over previous works. For example, on semantic segmentation task, with a similar amount of parameters, our LocalVim-S outperforms Vim-S by a large margin of 1.5 on mIoU (SS). 
\end{enumerate}

\section{Related Work}
\label{sec:related}

\subsection{Generic Vision Backbone Design}

The last decade has witnessed transformative advancements in computer vision, primarily driven by the evolution of deep neural networks and the emergence of foundational generic models. Initially, Convolutional Neural Networks (CNNs) \cite{krizhevsky2012imagenet,simonyan2014very,he2016deep,xie2017aggregated,sandler2018mobilenetv2,radosavovic2020designing,wang2018learning, han2022ghostnets} marked a significant milestone in visual model architecture, setting the stage for complex image recognition and analysis tasks. Among these works, ResNet \cite{he2016deep}, with a cornerstone residual connection technique, is one of the most popular model that is widely used in broad field of vision tasks; MobileNet \cite{howard2017mobilenets,sandler2018mobilenetv2} series lead the design of light-weight models with the utilization of depth-wise convolutions. However, the introduction of the Vision Transformer (ViT) \cite{dosovitskiy2021an} marked a paradigm shift, challenging the supremacy of CNNs in the domain. ViTs revolutionize the approach to image processing by segmenting images into a series of sequential patches and leveraging the self-attention mechanism, a core component of Transformer architectures \cite{vaswani2017attention}, to extract features. This novel methodology highlighted the untapped potential of Transformers in visual tasks, sparking a surge of research aimed at refining their architecture design\cite{su2022vitas} and training methodologies \cite{touvron2021training,liu2022tokenmix,touvron2022deit,he2022masked,xie2022simmim}, boosting computational efficiency \cite{wang2021pyramid,liu2021swin,chu2021twins,huang2022lightvit}, and extending their application scope \cite{strudel2021segmenter,lee2021vitgan,fang2021you,yang2022temporally,zhang2022styleswin,li2022exploring,peebles2023scalable}. Building on the success of long-sequence modeling with Mamba \cite{gu2023mamba}, a variant of State Space Models (SSMs), some innovative models such as Vim \cite{zhu2024vision} and VMamba \cite{liu2024vmamba} have been introduced in visual tasks, namely Vision Mamba. These models adapt the Mamba framework to serve as a versatile backbone for vision applications, demonstrating superior efficiency and accuracy over traditional CNNs and ViTs in high-resolution images.

\subsection{State Space Models}
State Space Models (SSMs) \cite{gu2021combining, gu2021efficiently, gupta2022diagonal, li2022makes, orvieto2023resurrecting}, represent a paradigm in architecture designed for sequence-to-sequence transformation, adept at managing long dependency tokens. Despite initial challenges in training, owing to their computational and memory intensity, recent advancements \cite{gu2021efficiently, gupta2022diagonal, gu2022parameterization, gu2023mamba, smith2022simplified} have significantly ameliorated these issues, positioning deep SSMs as formidable competitors against CNNs and Transformers. Particularly, S4 \cite{gu2021efficiently} introduced an efficient Normal Plus Low-Rank (NPLR) representation, leveraging the Woodbury identity for expedited matrix inversion, thus streamlining the convolution kernel computation. Building on this, Mamba \cite{gu2023mamba} further refined SSMs by incorporating an input-specific parameterization alongside a scalable, hardware-optimized computation approach, achieving unprecedented efficiency and simplicity in processing extensive sequences across languages and genomics.

The advent of S4ND \cite{nguyen2022s4nd} marked the initial foray of SSM blocks into visual tasks, adeptly handling visual data as continuous signals across 1D, 2D, and 3D domains. Subsequently, taking inspiration of the success of Mamba models, Vmamba \cite{liu2024vmamba} and Vim \cite{zhu2024vision} expanded into generic vision tasks, addressing the directional sensitivity challenge in SSMs by proposing bi-directional scan and cross-scan mechanisms. Leveraging Mamba's foundation in generic models, new methodologies have been developed for visual tasks, such as image segmentation \cite{ruan2024vm,liu2024swin,xing2024segmamba,ma2024u} and image synthetic \cite{guo2024mambair}, showcasing the adaptability and effectiveness of visual Mamba models in addressing complex vision challenges.

\section{Preliminaries}
\subsection{State Space Models}

Structured State Space Models (SSMs) represent a class of sequence models within deep learning, characterized by their ability to map a one-dimensional sequence $x(t) \in \mathbb{R}^{L}$ to $y(t) \in \mathbb{R}^{L}$ via an intermediate latent state $h(t) \in \mathbb{R}^{N}$:
\begin{align}\label{eq:ode}
    \begin{split}
        h'(t) &= \bm{A}h(t) + \bm{B}x(t),\\
        y(t) &= \bm{C}h(t),
    \end{split}
\end{align}
where the system matrices $\bm{A}\in\mathbb{R}^{N\times N}$, $\bm{B}\in\mathbb{R}^{N\times 1}$, and $\bm{C}\in\mathbb{R}^{N\times 1}$ govern the dynamics and output mapping, respectively.

\textbf{Discretization.} For practical implementation, the continuous system described by Equation \ref{eq:ode} is discretized using a zero-order hold assumption\footnote{This assumption holds the value of $x$ constant over a sample interval $\Delta$.}, effectively converting continuous-time parameters ($\bm{A}$, $\bm{B}$) to their discrete counterparts ($\bm{\overline{A}}$, $\bm{\overline{B}}$) over a specified sampling timescale $\bm{\Delta} \in \mathbb{R}{>0}$:
\begin{align}
\begin{split}
    \bm{\overline{A}} &= e^{\bm{\Delta}\bm{A}} \\
    \bm{\overline{B}} &= (\bm{\Delta}\bm{A})^{-1} (e^{\bm{\Delta}\bm{A}} - \bm{I}) \cdot \bm{\Delta}\bm{B}.
\end{split}
\end{align}
This leads to a discretized model formulation as follows:
\begin{align}\label{eq:discretization}
    \begin{split}
        h_t &= \bm{\overline{A}}h_{t-1} + \bm{\overline{B}}x_t,\\
        y_t &= \bm{C}h_t.
    \end{split}
\end{align}

For computational efficiency, the iterative process delineated in Equation \ref{eq:discretization} can be expedited through parallel computation, employing a global convolution operation:
\begin{align}
    \begin{split}
        \bm{y} &= \bm{x} \circledast \bm{\overline{K}} \\
        \text{with} \quad \bm{\overline{K}} &= (\bm{C}\bm{\overline{B}},\bm{C}\overline{\bm{A}\bm{B}}, ..., \bm{C}\bm{\overline{A}}^{L-1}\bm{\overline{B}}),
    \end{split}
\end{align}
where $\circledast$ represents the convolution operation, and $\bm{\overline{K}} \in \mathbb{R}^L$ serves as the kernel of the SSM. This approach leverages the convolution to synthesize the outputs across the sequence simultaneously, enhancing computational efficiency and scalability.

\subsection{Selective State Space Models}

Traditional State Space Models (SSMs), often referred to as S4, have achieved linear time complexity. However, their ability to capture sequence context is inherently constrained by static parameterization. To address this limitation, Selective State Space Models (termed Mamba) \cite{gu2023mamba} introduce a dynamic and selective mechanism for the interactions between sequential states. Unlike conventional SSMs that utilize constant transition parameters $(\bm{\overline{A}}, \bm{\overline{B}})$, Mamba models employ input-dependent parameters, enabling a richer, sequence-aware parameterization. Specifically, Mamba models calculate the parameters $\bm{B} \in \mathbb{R}^{B\times L\times N}$, $\bm{C} \in \mathbb{R}^{B\times L\times N}$, and $\bm{\Delta} \in \mathbb{R}^{B\times L\times D}$ directly from the input sequence $\bm{x} \in \mathbb{R}^{B\times L\times D}$. 

The Mamba models, leveraging selective SSMs, not only achieve linear scalability in sequence length but also deliver competitive performance in language modeling tasks. This success has inspired subsequent applications in vision tasks, with studies proposing the integration of Mamba into foundational vision models. Vim \cite{zhu2024vision}, adopts a ViT-like architecture, incorporating bi-directional Mamba blocks in lieu of traditional transformer blocks. VMamba \cite{liu2024vmamba} introduces a novel 2D selective scanning technique to scan images in both horizontal and vertical orientations, and constructs a hierarchical model akin to the Swin Transformer \cite{liu2021swin}. Our research extends these initial explorations, focusing on optimizing the S6 adaptation for vision tasks, where we achieve improved performance outcomes.

\begin{figure}[tb]
  \centering
  \hspace{2mm}
  \begin{subfigure}{0.6\linewidth}
    \includegraphics[width=\linewidth]{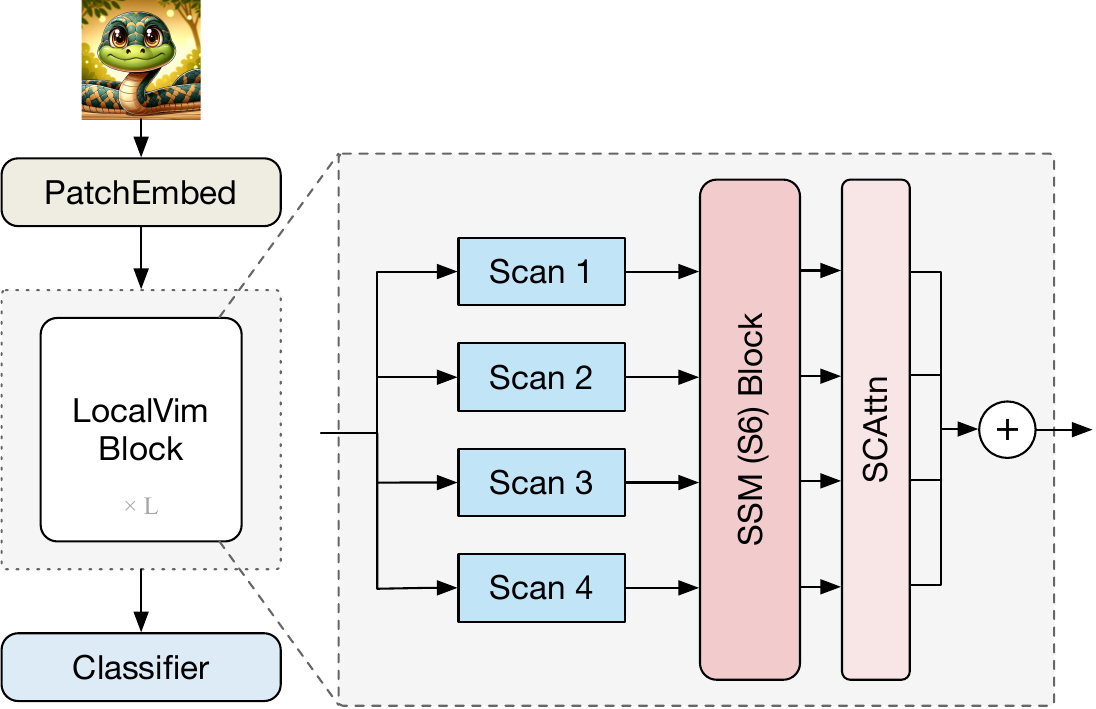}
    \caption{}
    \label{fig:structure}
  \end{subfigure}
  \hfill
  \begin{subfigure}{0.3\linewidth}
    \includegraphics[width=\linewidth]{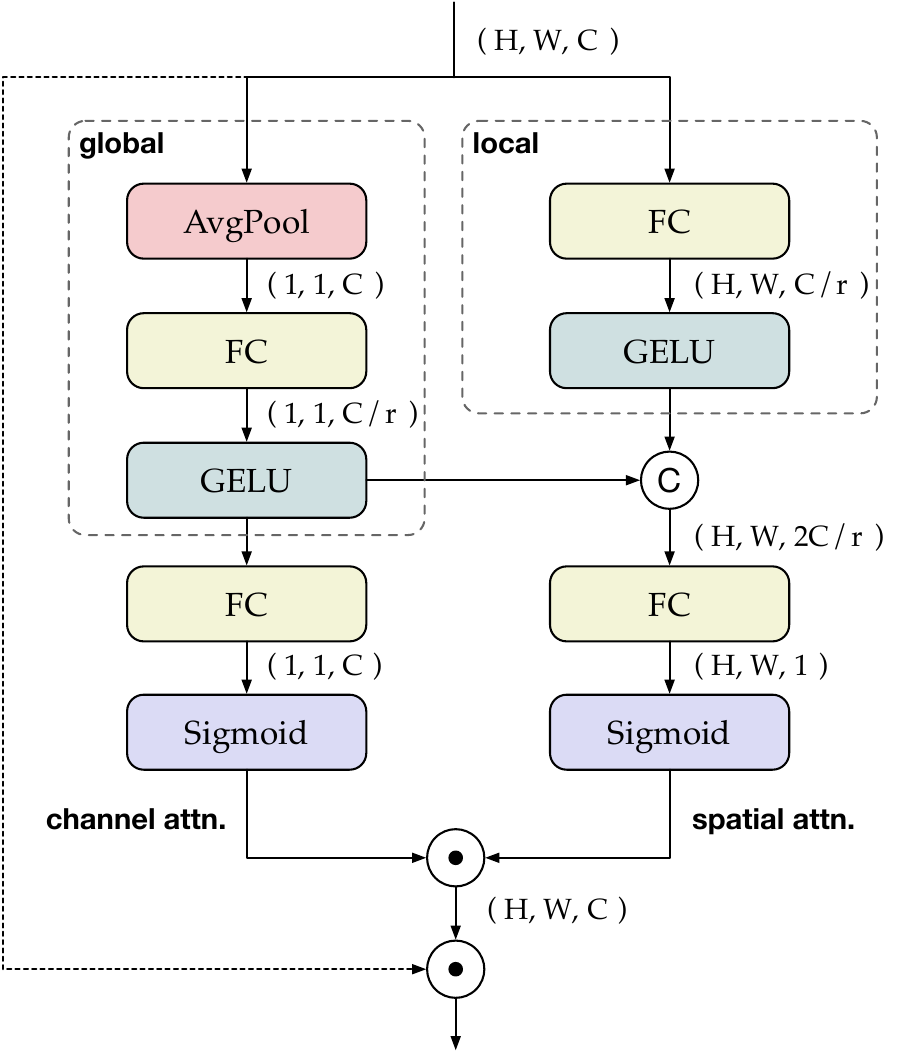}
    \caption{}
    \label{fig:scattn}
  \end{subfigure}
  \hspace{2mm}
  \caption{(a) Structure of the LocalVim model. (b) Illustration of the proposed spatial and channel attention module (SCAttn).}
  \label{fig:arch}
\end{figure}

\section{Methodology}

This section delineates core components of our LocalMamba, beginning with the local scan mechanism designed to enhance the model's ability to dig fine-grained details from images. Subsequently, we introduce the scan direction search algorithm, an innovative approach that identifies optimal scanning sequences across different layers, thereby ensuring a harmonious integration of global and local visual cues. The final part of this section illustrates the deployment of the LocalMamba framework within both simple plain architecture and complex hierarchical architecture, showcasing its versatility and effectiveness in diverse settings.

\subsection{Local Scan for Visual Representations} \label{sec:local_scan}

Our method employs the selective scan mechanism, S6, which has shown exceptional performance in handling 1D causal sequential data. This mechanism processes inputs causally, effectively capturing vital information within the scanned segments, akin to language modeling where understanding the dependencies between sequential words is essential. However, the inherent non-causal nature of 2D spatial data in images poses a significant challenge to this causal processing approach. Traditional strategies that flatten spatial tokens compromise the integrity of local 2D dependencies, thereby diminishing the model’s capacity to effectively discern spatial relationships. For instance, as depicted in Figure \ref{fig:scan_illustration} (a) and (b), the flattening approach utilized in Vim \cite{zhu2024vision} disrupts these local dependencies, significantly increasing the distance between vertically adjacent tokens and hampering the model's ability to capture local nuances. While VMamba \cite{liu2024vmamba} attempts to address this by scanning images in both horizontal and vertical directions, it still falls short of comprehensively processing the spatial regions in a single scan.

To address this limitation, we introduce a novel approach for scanning images locally. By dividing images into multiple distinct local windows, our method ensures a closer arrangement of relevant local tokens, enhancing the capture of local dependencies. This technique is depicted in Figure \ref{fig:scan_illustration} (c), contrasting our approach with prior methods that fail to preserve spatial coherence.

While our method excels at capturing local dependencies effectively within each region, it also acknowledges the significance of global context. To this end, we construct our foundational block by integrating the selective scan mechanism across four directions: the original (a) and (c) directions, along with their flipped counterparts facilitating scanning from tail to head (the flipped directions are adopted in both Vim and VMamba for better modeling of non-causual image tokens). This multifaceted approach ensures a comprehensive analysis within each selective scan block, striking a balance between local detail and global perspective.

As illustrated in Figure \ref{fig:arch}, our block processes each input image feature through four distinct selective scan branches. These branches independently capture relevant information, which is subsequently merged into a unified feature output. To enhance the integration of diverse features and eliminate extraneous information, we introduce a spatial and channel attention module before merging. As shown in Figure \ref{fig:scattn}, this module adaptively weights the channels and tokens within the features of each branch, comprising two key components: a channel attention branch and a spatial attention branch. The channel attention branch aggregates global representations by averaging the input features across the spatial dimension, subsequently applying a linear transformation to determine channel weights. Conversely, the spatial attention mechanism assesses token-wise significance by augmenting each token's features with global representations, enabling a nuanced, importance-weighted feature extraction.

\textbf{Remark.} While some ViT variants, such as the Swin Transformer \cite{liu2021swin}, propose the division of images into smaller windows, the local scan in our LocalMamba is distinct both in purpose and effect. The windowed self-attention in ViTs primarily addresses the computational efficiency of global self-attention, albeit at the expense of some global attention capabilities. Conversely, our local scan mechanism aims to rearrange token positions to enhance the modelling of local region dependencies in visual Mamba, while the global understanding capability is retained as the entire image is still aggregated and processed by SSM.

\subsection{Searching for Adaptive Scan}

The efficacy of the Structured State Space Model (SSM) in capturing image representations varies across different scan directions. Achieving optimal performance intuitively suggests employing multiple scans across various directions, similar to our previously discussed 4-branch local selective scan block. However, this approach substantially increases computational demands. To address this, we introduce a strategy to efficiently select the most suitable scan directions for each layer, thereby optimizing performance without incurring excessive computational costs. This method involves searching for the optimal scanning configurations for each layer, ensuring a tailored and efficient representation modeling.

\begin{figure}[t]
    \centering
    \includegraphics[width=\linewidth]{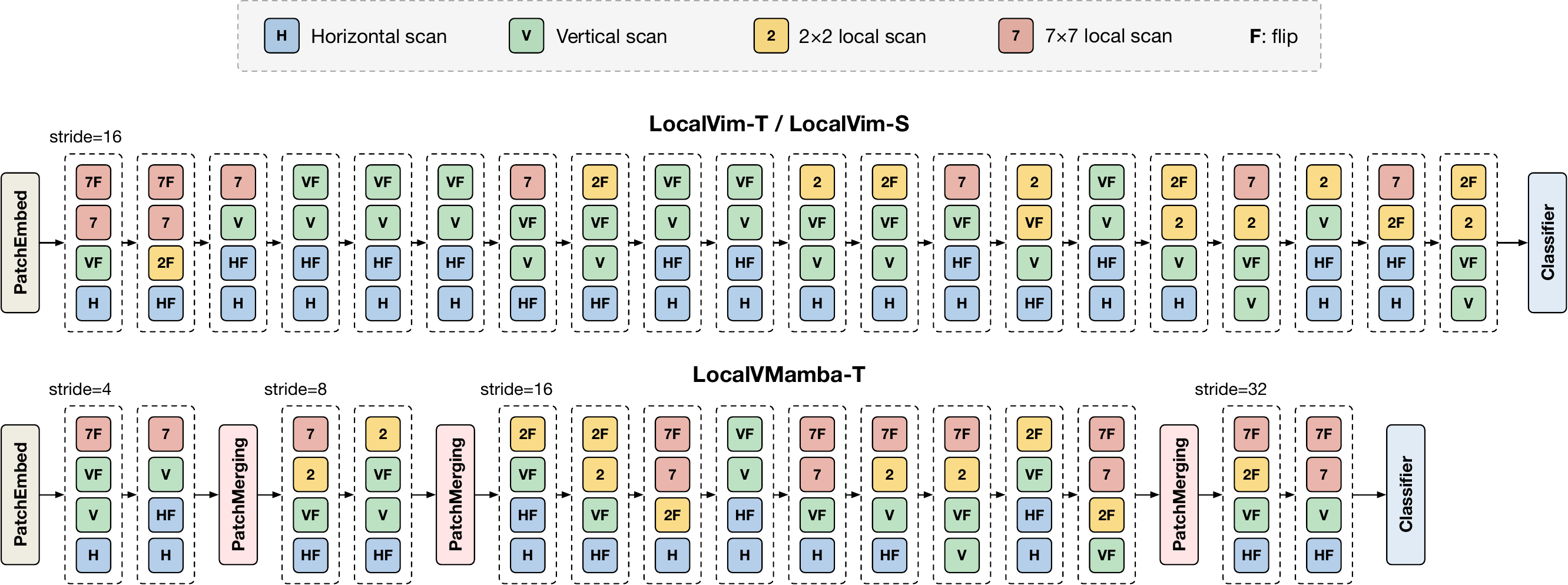}
    \caption{Visualization of the searched directions of our models. The visualization of LocalVMamba-S is in Section \ref{sec:vis_searched_s}.}
    \label{fig:vis_searched}
\end{figure}

\textbf{Search space.} To tailor the scanning process for each layer, we introduce a diverse set $\mathcal{S}$ of $8$ candidate scan directions. These include horizontal and vertical scans (both standard and flipped), alongside local scans with window sizes of $2$ and $7$ (also both standard and flipped). For a consistent computational budget as previous models, we select $4$ out of these $8$ directions for each layer. This approach results in a substantial search space of $(C_8^4)^K$, with $K$ representing the total number of blocks.

Building upon the principles of DARTS \cite{liu2018darts}, our method applies a differentiable search mechanism for scan directions, employing continuous relaxation to navigate the categorical choices. This approach transforms the discrete selection process into a continuous domain, allowing for the use of softmax probabilities to represent the selection of scan directions:
\begin{equation}
    \bm{y}^{(l)} = \sum_{s\in\mathcal{S}} \frac{\mathrm{exp}(\alpha_s^{(l)})}{\sum_{s'\in\mathcal{S}} \mathrm{exp}( \alpha_{s'}^{(l)})} \mathrm{SSM}_s( \bm{x}^{(l)}),
\end{equation}
where $\bm{\alpha}^{(l)}$ denotes a set of learnable parameters for each layer $l$, reflecting the softmax probabilities over all potential scan directions.

We construct the entire search space as an over-parameterized network, allowing us to simultaneously optimize the network parameters and the architecture variables $\bm{\alpha}$, following standard training protocols. Upon completion of the training, we derive the optimal direction options by selecting the four directions with the highest softmax probabilities. We visualize the searched directions of our models in Figure \ref{fig:vis_searched}. For detailed analysis of the search results, see Section \ref{sec:vis_searched}.

\textbf{Scalability of direction search.} Our current approach aggregates all scan directions for selection in training, aptly serving models with a moderate range of options. For instance, a model featuring $20$ blocks and $128$ directions per block requires $28$ GB of GPU memory, indicating scalability limits for extensive choices. To mitigate memory consumption in scenarios with a vast selection array, techniques such as single-path sampling \cite{guo2020single,you2020greedynas}, binary approximation \cite{cai2018proxylessnas}, and partial-channel usage \cite{xu2020pcdarts} present viable solutions. We leave the investigation of more adaptive direction strategies and advanced search techniques to future endeavors.

\input{tables/archs.tex}

\subsection{Architecture Variants}

To thoroughly assess our methodology's effectiveness, we introduce architecture variants grounded in both plain \cite{zhu2024vision} and hierarchical \cite{liu2024vmamba} structures, named LocalVim and LocalVMamba, respectively. The configurations of these architectures are detailed in Table \ref{tab:archs}. Specifically, in LocalVim, the standard SSM block is substituted with our LocalVim block, as depicted in Figure \ref{fig:arch}. Considering the original Vim block comprises two scanning directions (horizontal and flipped horizontal), and our LocalVim introduces four scanning directions, thereby increasing the computational overhead. To maintain similar computation budgets, we adjust the number of Vim blocks from 24 to 20. For LocalVMamba, which inherently has four scanning directions akin to our model, we directly replace the blocks without changing the structural configurations.

\textbf{Computational cost analysis.} Our LocalMamba block is efficient and effective, with only a marginal increase on computation cost. The scanning mechanism, which involves merely repositioning tokens, incurs no additional computational cost in terms of FLOPs. Furthermore, the SCAttn module, designed for efficient aggregation of varied information across scans, is exceptionally streamlined. It leverages linear layers to reduce the token dimension by a factor of $1/r$, thereafter generating attention weights across both spatial and channel dimensions, with $r$ set to 8 for all models. For instance, our LocalVMamba-T model, which replaces the VMamba block with our LocalMamba block, only increases the FLOPs of VMamva-T from 5.6G to 5.7G.

\section{Experiments}

This section outlines our experimental evaluation, starting with the ImageNet classification task, followed by transferring the trained models to various downstream tasks, including object detection and semantic segmentation.

\subsection{ImageNet Classification}

\textbf{Training strategies.} We train the models on ImageNet-1K dataset \cite{Imagenet} and evaluate the performance on ImageNet-1K validation set. Following previous works \cite{touvron2021training,liu2021swin,zhu2024vision,liu2024vmamba}, we train our models for $300$ epochs with a base batch size of $1024$ and an AdamW optimizer, a cosine annealing learning rate schedule is adopted with initial value $10^{-3}$ and 20-epoch warmup. For training data augmentation, we use random cropping, AutoAugment \cite{cubuk2019autoaugment} with policy \textit{rand-m9-mstd0.5}, and random erasing of pixels with a probability of $0.25$ on each image, then a MixUp strategy with ratio $0.2$ is adopted in each batch. An exponential moving average on model is adopted with decay rate $0.9999$.

\input{tables/imagenet.tex}

\textbf{Scan direction search.} For the supernet training, we curtail the number of epochs to $100$ while maintaining the other hyper-parameters consistent with standard ImageNet training. The embedding dimension for the supernet in LocalVim variants is set to $128$, with search operations conducted identically on LocalVim-T and LocalVim-S due to their uniform layer structure. For LocalVMamba variants, including LocalVMamba-T and LocalVMamba-S, the initial embedding dimension is minimized to $32$ to facilitate the search process.

\textbf{Results.} Our results, summarized in Table \ref{tab:imagenet}, illustrate significant accuracy enhancements over traditional CNNs and ViT methodologies. Notably, LocalVim-T achieves a $76.2\%$ accuracy rate with 1.5G FLOPs, surpassing the DeiT-Ti, which records a $72.2\%$ accuracy. In hierarchical structures, LocalVMamba-T's $82.7\%$ accuracy outperforms Swin-T by $1.4\%$. Moreover, compared to our seminal contributions, Vim and VMamba, our approach registers substantial gains; for instance, LocalVim-T and LocalVMamba-T exceed Vim-Ti and VMamba-T by $2.7\%$ and $0.5\%$ in accuracy, respectively. Additionally, to validate the local scan's effectiveness, we conducted additional experiments on models without the scan direction search, delineated in Section \ref{sec:local_scan}, marked with $^*$ in the table. Incorporating merely our local scans into the original Vim framework, LocalVim-T$^*$ surpasses Vim-Ti by $2.7\%$, while the complete methodology further elevates accuracy by $0.4\%$. These findings affirm the pivotal role of scan directions in visual SSMs, evidencing our local scan approach's capability to enhance local dependency capture effectively.

\input{tables/coco.tex}

\subsection{Object Detection}
\textbf{Training strategies.}  We validate our performance on object detection using MSCOCO 2017 dataset \cite{lin2014microsoft} and MMDetection library \cite{chen2019mmdetection}. For LocalVMamba series, we follow previous works \cite{liu2024vmamba,liu2021swin} to train object detection and instance segmentation tasks with Mask-RCNN detector \cite{he2017mask}. The training strategies include $1\times$ setting of $12$ training epochs and $3\times$ setting with $36$ training epochs and multi-scale data augmentations. While for LocalVim, we follow Vim \cite{zhu2024vision} to use Cascade Mask R-CNN with ViTDet \cite{li2022exploring} as the detector.

\textbf{Results.} We summarize our results on LocalVMamba in comparisons to other backbones in Table \ref{tab:coco}. We can see that, our LocalVMamba outperforms VMamba consistently on all the model variants. And compared to other architectures, CNNs and ViTs, we obtain significant superiority. For example, our LocalVMamba-T obtains 46.7 box AP and 42.2 mask AP, improves Swin-T by large margins of 4.0 and 2.9, respectively. For quantitative comparisons to Vim, please refer to supplementary material.

\input{tables/ADE20K.tex}

\subsection{Semantic Segmentation}

\textbf{Training strategies.} Following \cite{liu2021swin,liu2024vmamba,zhu2024vision}, we train UperNet \cite{xiao2018unified} with our backbones on ADE20K \cite{zhou2019semantic} dataset. The models are trained with a total batch size of $16$ with $512\times 512$ inputs, an AdamW optimizer is adopted with weight decay 0.01. We use a Poly learning rate schedule, which decays 160K iterations with an initial learning rate of $6\times10^{-5}$. Note that Vim did not report the FLOPs and mIoU (MS) and release the code for segmentation, so we implement our LocalVim following the ViT example configuration in MMSegmentation \cite{mmseg2020}.

\textbf{Results.} We report the results of both LocalVim and LocalVMamba in Table \ref{tab:ADE20K}. On LocalVim, we achieve significant improvements over the baseline Vim-Ti. For example, with a similar amount of parameters, our LocalVim-S outperforms Vim-S by 1.5 on mIoU (SS). While on LocalVMamba, we achieve significant improvements over the VMamba baseline; \eg, our LocalVMamba-T achieves a remarkable mIoU (MS) of 49.1, surpassing VMamba-T by 0.8. Compared to CNNs and ViTs, our improvements are more obvious. The results demonstrate the efficacy of the global representation of SSMs in dense prediction tasks.

\subsection{Ablation Study}

\textbf{Effect of local scan.} The impact of our local scan technique is assessed, with experiments detailed in Table \ref{tab:ab_scan}. Substituting Vim-T's traditional horizontal scan with our local scan yielded a 1\% performance boost over the baseline. A combination of scan directions under a constrained FLOP budget in LocalVim-T$^*$ led to an additional 1.1\% accuracy increase. These results underscore the varied impacts of scanning with different window sizes (considering the horizontal scan as a local scan with a window size of $14\times14$) on image recognition, and an amalgamation of these scans enhances performance further.

\textbf{Effect of SCAttn.} In Table \ref{tab:ab_scan}, the incorporation of SCAttn into the final LocalVim block facilitated an additional improvement of 0.6\%, validating the effectiveness of strategically combining various scan directions. This underscores SCAttn's role in enhancing performance by adaptively merging scan directions.

\input{tables/ab_scan.tex}

\textbf{Effect of scan direction search.} Our empirical evaluation, as depicted in Table \ref{tab:imagenet}, confirms the significant benefits derived from the scan direction search strategy in the final LocalVim models. These models exhibit marked improvements over versions that merely amalgamate horizontal scans, local scans with a window size of $2\times2$, and their mirrored counterparts. For instance, LocalVim-T exhibits a $0.4\%$ enhancement over LocalVim-T$^*$. This performance gain can be attributed to the methodological selection of scan combinations at each layer, offering a diverse set of options to optimize model efficacy.

\subsection{Visualization of Searched Scan Directions} \label{sec:vis_searched}
Figure \ref{fig:vis_searched} presents visualizations of the scanned directions obtained in our models. Observations suggest that within the plain architecture of LocalVim, there is a predilection for employing local scans in both the initial and terminal segments, with intermediate layers favoring global horizontal and vertical scans. Notably, the $2\times2$ local scans tend to concentrate towards the network's tail, whereas larger $7\times7$ scans are prominent towards the network's inception. Conversely, the hierarchical structure of LocalVMamba exhibits a greater inclination towards local scans compared to LocalVim, with a preference for $7\times7$ scans over $2\times2$ scans.

\section{Conclusion}
In this paper, we introduce LocalMamba, an innovative approach to visual state space models that significantly enhances the capture of local dependencies within images while maintaining global contextual understanding. Our method leverages windowed selective scanning and scan direction search to significantly improve upon existing models. Extensive experiments across various datasets and tasks have demonstrated the superiority of LocalMamba over traditional CNNs and ViTs, establishing new benchmarks for image classification, object detection, and semantic segmentation. Our findings underscore the importance of scanning mechanisms in visual state space model and open new avenues for research in efficient and effective state space modeling. Future work will explore the scalability of our approach to more complex and diverse visual tasks, as well as the potential integration of more advanced scanning strategies.

%
%
\bibliographystyle{splncs04}
\bibliography{main}

\newpage
\appendix
\section{Appendix}
\subsection{Comparison to Vim on Object Detection} \label{sec:cmp_det_vim}

Different from VMamba that uses the Mask R-CNN framework for benchmark, Vim utilizes the neck architecture in ViTDet and train Cascade Mask R-CNN as the detector. We also align with the settings in Vim for fair comparison and evaluation of our LocalVim model.

We summarize the results in Table \ref{tab:coco_vim}. Our LocalVim-T performs on pair with the Vim-Ti, while has significant superiorities in AP$^{b}_{50}$ and mask AP. For example, our LocalVim-T improves Vim-Ti on AP$^{m}$ and  AP$^{m}_{50}$ by 0.7 and 2.1, respectively.

\input{tables/coco_vim}

\begin{figure}[h]
    \centering
    \includegraphics[width=\linewidth]{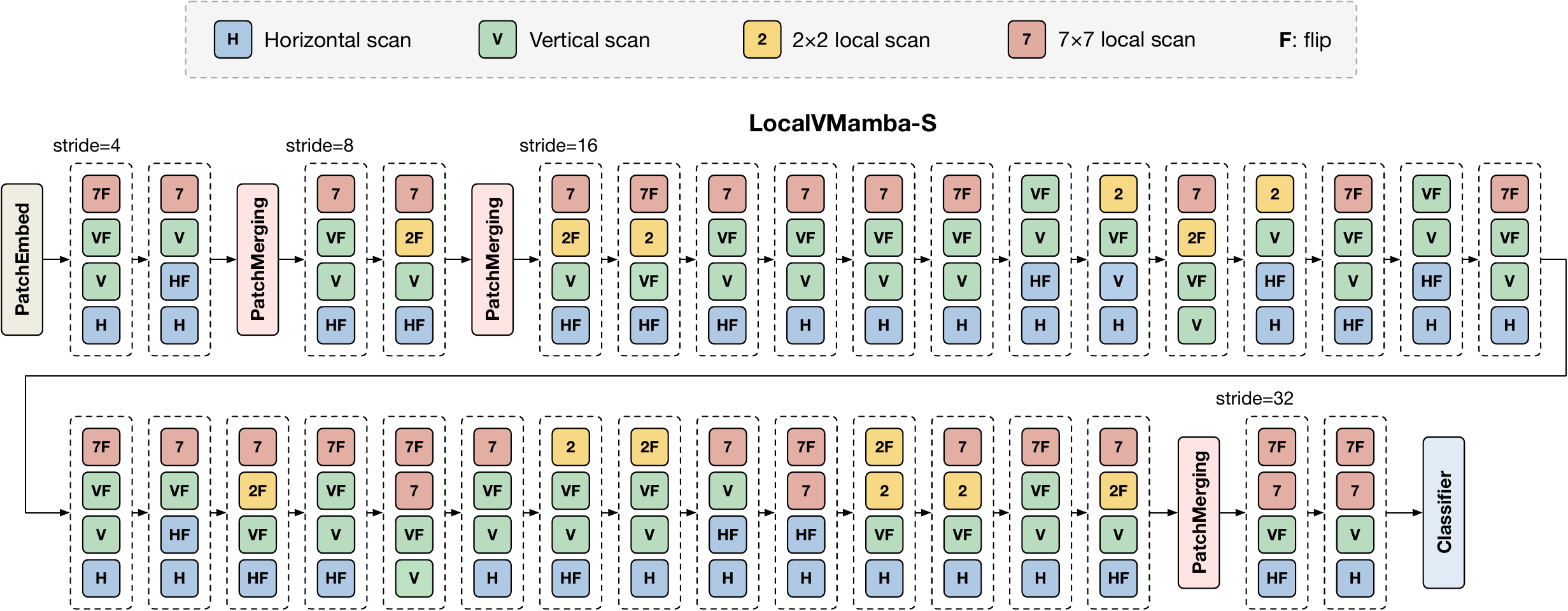}
    \caption{Visualization of the searched directions of LocalVMamba-S.}
    \label{fig:vis_searched_s}
\end{figure}

\subsection{Visualization of Searched Directions on LocalVMamba-S} \label{sec:vis_searched_s}

We visualize the searched directions of LocalVMamba-S in Figure \ref{fig:vis_searched_s}. In this model, with $27$ layers in stage 3, more $7\times 7$ local scans are preferred compared to LocalVMamba-T.

\subsection{Discussions}
\textbf{Potential negative impact.} Investigating the effects of the proposed model requires large consumptions on computation resources, which can potentially raise the environmental concerns. 

\textbf{Limitations.} The visual state space models with linear-time complexity to the sequence length, show significant improvements especially on large-resolution downstream tasks compared to previous CNNs and ViTs architectures. Nonetheless, the computational framework of SSMs is inherently more intricate than that of convolution and self-attention mechanisms, complicating the efficient execution of parallel computations. Current deep learning frameworks also exhibit limited capability in accelerating SSM computations as efficiently as they do for more established architectures. On a positive note, ongoing efforts in projects such as VMamba\footnote{Project page: https://github.com/MzeroMiko/VMamba.} \cite{liu2024vmamba} are aimed at enhancing the computational efficiency of selective SSM operations. These initiatives have already realized notable advancements in speed, as evidenced by the improvements over the original implementations documented in Mamba \cite{gu2023mamba}.

\end{document}

%% file: lib/my_instructions.tex
\input{lib/math_commands.tex}

\definecolor{mycyan}{RGB}{212, 239, 251}
\usepackage{colortbl}
\usepackage{xcolor}
\definecolor{mygray}{gray}{.9}
\definecolor{goldenrod}{RGB}{245,245,220}
\newlength\savewidth
\newcolumntype{a}{>{\columncolor{mygray}}c}
\usepackage{fontawesome5}
\usepackage{booktabs}
\usepackage{makecell}
\usepackage{fontawesome5}
\usepackage{lipsum}
\usepackage{comment}
\usepackage{multirow}
\usepackage{wrapfig}
\usepackage{algorithm}
\usepackage{algorithmic}
\usepackage{xfrac}  

\usepackage{graphicx}
\usepackage{color}

\definecolor{darkgreen}{rgb}{0,0.7,0}
\newcommand{\blue}[1]{{\color{blue}{#1}}}

\definecolor{mygraytext}{gray}{.5}

\def\eg{\emph{e.g.}}


%% file: lib/math_commands.tex
\usepackage{amsmath,amsfonts,bm}

\def\eqref#1{equation~\ref{#1}}

\def\1{\bm{1}}

\DeclareMathAlphabet{\mathsfit}{\encodingdefault}{\sfdefault}{m}{sl}
\SetMathAlphabet{\mathsfit}{bold}{\encodingdefault}{\sfdefault}{bx}{n}



%% file: tables/archs.tex
\begin{table}[t]
    \centering
    \setlength{\tabcolsep}{1.5mm}
    \renewcommand{\arraystretch}{1.1}
    \caption{Architecture variants. We follow the original structure designs of Vim and VMamba, where Vim uses a plain structure with a patch embedding of stride $16$, while VMamba constructs a hierarchical structures with SSM stages on strides 4, 8, 16, and 32.}
    \small
    \begin{tabular}{l|c|c|c|c}
    \toprule
    Model & \#Dims & \#Blocks & Params & FLOPs\\
    \midrule
    LocalVim-T & 192 & 20 & 8M & 1.5G\\
    LocalVim-S & 384 & 20 & 28M & 4.8G\\
    \midrule
    LocalVMamba-T & [96, 192, 384, 768] & [2, 2, 9, 2] & 26M & 5.7G\\
    LocalVMamba-S & [96, 192, 384, 768] & [2, 2, 27, 2] & 50M & 11.4G\\
    \bottomrule
    \end{tabular}
    \label{tab:archs}
\end{table}

%% file: tables/imagenet.tex
\begin{table}[t]
    \centering
    \setlength{\tabcolsep}{2mm}
    \renewcommand{\arraystretch}{1.1}
    \caption{Comparison of different backbones on ImageNet-1K classification. $*$: Our model without scan direction search.}
    \small
    \begin{tabular}{l|ccc|c}
    \toprule
    Method & Image size & Params (M) & FLOPs (G) & Top-1 ACC (\%)\\
    \midrule
    RegNetY-4G \cite{radosavovic2020designing} & $224^2$ & 21 & 4.0 & 80.0\\
    RegNetY-8G \cite{radosavovic2020designing} & $224^2$ & 39 & 8.0 & 81.7\\
    RegNetY-16G \cite{radosavovic2020designing} & $224^2$ & 84 & 16.0 & 82.9\\
    \midrule
    ViT-B/16 \cite{dosovitskiy2021an} & $384^2$ & 86 & 55.4 & 77.9\\
    ViT-L/16 \cite{dosovitskiy2021an} & $384^2$ & 307 & 190.7 & 76.5\\
    \midrule
    DeiT-Ti \cite{touvron2021training} & $224^2$ & 6 & 1.3 & 72.2\\
    DeiT-S \cite{touvron2021training} & $224^2$ & 22 & 4.6 & 79.8\\
    DeiT-B \cite{touvron2021training} & $224^2$ & 86 & 17.5 & 81.8\\
    \midrule
    Swin-T \cite{liu2021swin} & $224^2$ & 29 & 4.5 & 81.3\\
    Swin-S \cite{liu2021swin} & $224^2$ & 50 & 8.7 & 83.0\\
    Swin-B \cite{liu2021swin} & $224^2$ & 88 & 15.4 & 83.5\\
    \midrule
    Vim-Ti \cite{zhu2024vision} & $224^2$ & 7 & 1.5 & 73.1\\
    Vim-S \cite{zhu2024vision} & $224^2$ & 26 & 5.1 & 80.3\\
    \rowcolor{mygray}
    LocalVim-T$^*$ & $224^2$ & 8 & 1.5 & 75.8\\
    \rowcolor{mygray}
    LocalVim-T & $224^2$ & 8 & 1.5 & 76.2\\
    \rowcolor{mygray}
    LocalVim-S$^*$ & $224^2$ & 28 & 4.8 & 81.0\\
    \rowcolor{mygray}
    LocalVim-S & $224^2$ & 28 & 4.8 & 81.2\\
    \midrule
    VMamba-T \cite{liu2024vmamba} & $224^2$ & 22 & 5.6 & 82.2\\
    VMamba-S \cite{liu2024vmamba} & $224^2$ & 44 & 11.2 & 83.5\\
    VMamba-B \cite{liu2024vmamba} & $224^2$ & 75 & 18 & 83.7\\
    \rowcolor{mygray}
    LocalVMamba-T & $224^2$ & 26 & 5.7 & 82.7\\
    \rowcolor{mygray}
    LocalVMamba-S & $224^2$ & 50 & 11.4 & 83.7 \\
    \bottomrule
    \end{tabular}
    \label{tab:imagenet}
\end{table}

%% file: tables/coco.tex
\begin{table}[t]
    \centering
    \renewcommand{\arraystretch}{1.1}
    \setlength{\tabcolsep}{1.5mm}
    \caption{Object detection and instance segmentation results on COCO \textit{val} set.}
    \small
    \begin{tabular}{l|cc|ccc|ccc}
    \toprule
    \multicolumn{9}{c}{\textbf{Mask R-CNN 1$\times$ schedule}}\\
    \midrule
    Backbone & Params & FLOPs & AP$^\mathrm{b}$ & AP${}^\mathrm{b}_{50}$ & AP${}^\mathrm{b}_{75}$ & AP${}^\mathrm{m}$ & AP${}^\mathrm{m}_{50}$ & AP${}^\mathrm{m}_{75}$\\
    \midrule
    ResNet-50 & 44M & 260G & 38.2 & 58.8 & 41.4 & 34.7 & 55.7 & 37.2\\
    Swin-T & 48M & 267G & 42.7 & 65.2 & 46.8 & 39.3 & 62.2 & 42.2 \\
    ConvNeXt-T & 48M & 262G & 44.2 & 66.6 & 48.3 & 40.1 & 63.3 & 42.8\\
    ViT-Adapter-S & 48M & 403G & 44.7 & 65.8 & 48.3 & 39.9 & 62.5 & 42.8 \\
    VMamba-T & 42M & 286G & 46.5 & 68.5 & 50.7 & 42.1 & 65.5 & 45.3 \\
    \rowcolor{mygray}
    LocalVMamba-T & 45M & 291G & 46.7 & 68.7 & 50.8 & 42.2 & 65.7 & 45.5\\
    \midrule
    ResNet-101 & 63M & 336G & 38.2 & 58.8 & 41.4 & 34.7 & 55.7 & 37.2\\
    Swin-S & 69M & 354G & 44.8 & 66.6 & 48.9 & 40.9 & 63.2 & 44.2 \\
    ConvNeXt-S & 70M & 348G & 45.4 & 67.9 & 50.0 & 41.8 & 65.2 & 45.1 \\
    VMamba-S & 64M & 400G & 48.2 & 69.7 & 52.5 & 43.0 & 66.6 & 46.4 \\
    \rowcolor{mygray}
    LocalVMamba-S & 69N & 414G & 48.4 & 69.9 & 52.7 & 43.2 & 66.7 & 46.5\\
    \midrule
    \multicolumn{9}{c}{\textbf{Mask R-CNN 3$\times$ MS schedule}}\\
    \midrule
    Swin-T & 48M & 267G & 46.0 & 68.1 & 50.3 & 41.6 & 65.1 & 44.9 \\
    ConvNeXt-T & 48M & 262G & 46.2 & 67.9 & 50.8 & 41.7 & 65.0 & 44.9 \\
    ViT-Adapter-S & 48M & 403G & 48.2 & 69.7 & 52.5 & 42.8 & 66.4 & 45.9 \\
    VMamba-T & 42M & 286G & 48.5 & 69.9 & 52.9 & 43.2 & 66.8 & 46.3 \\
    \rowcolor{mygray}
    LocalVMamba-T & 45M & 291G & 48.7 & 70.1 & 53.0 & 43.4 & 67.0 & 46.4\\
    \midrule
    Swin-S & 69M & 354G & 48.2 & 69.8 & 52.8 & 43.2 & 67.0 & 46.1 \\
    ConvNeXt-S & 70M & 348G & 47.9 & 70.0 & 52.7 & 42.9 & 66.9 & 46.2 \\
    VMamba-S & 64M & 400G & 49.7 & 70.4 & 54.2 & 44.0 & 67.6 & 47.3 \\
    \rowcolor{mygray}
    LocalVMamba-S & 69M & 414G & 49.9 & 70.5 & 54.4 & 44.1 & 67.8 & 47.4\\
    \bottomrule
    \end{tabular}
    \label{tab:coco}
\end{table}

%% file: tables/ADE20K.tex
\begin{table}[t]
    \centering
    \setlength{\tabcolsep}{1mm}
    \renewcommand{\arraystretch}{1.1}
    \caption{Results of semantic segmentation on ADE20K using UperNet \cite{xiao2018unified}. We measure the mIoU with single-scale (SS) and multi-scale (MS) testings on the \textit{val} set. The FLOPs are measured with an input size of $512\times2048$. -: Vim~\cite{zhu2024vision} did not report the FLOPs and mIOU (MS). MLN: multi-level neck.}
    \small
    \begin{tabular}{l|ccc|cc}
    \toprule
    Backbone & Image size & Params (M) & FLOPs (G) & mIoU (SS) & mIoU (MS)\\
    \midrule
    DeiT-Ti & $512^2$ & 11 & - & 39.2 & \blue{-}\\
    Vim-Ti & $512^2$ & 13 & - & 40.2 & -\\
    \rowcolor{mygray}
    LocalVim-T & $512^2$ & 36 & 181 & 43.4 & 44.4\\
    \midrule
    ResNet-50 & $512^2$ & 67 & 953 & 42.1 & 42.8\\
    DeiT-S + MLN & $512^2$ & 58 & 1217 & 43.8 & 45.1\\
    Swin-T & $512^2$ & 60 & 945 & 44.4 & 45.8\\
    Vim-S & $512^2$ & 46 & - & 44.9 & -\\
    \rowcolor{mygray}
    LocalVim-S & $512^2$ & 58 & 297 & 46.4 & 47.5\\
    VMamba-T & $512^2$ & 55 & 964 & 47.3 & 48.3\\
    \rowcolor{mygray}
    LocalVMamba-T & $512^2$ & 57 & 970 & 47.9 & 49.1\\
    \midrule
    ResNet-101 & $512^2$ & 85 & 1030 & 42.9 & 44.0\\
    DeiT-B + MLN & $512^2$ & 144 & 2007 & 45.5 & 47.2\\
    Swin-S & $512^2$ & 81 & 1039 & 47.6 & 49.5\\
    VMamba-S & $512^2$ & 76 & 1081 & 49.5 & 50.5\\
    \rowcolor{mygray}
    LocalVMamba-S & $512^2$ & 81 & 1095 & 50.0 & 51.0 \\
    \bottomrule
    \end{tabular}
    \label{tab:ADE20K}
\end{table}

%% file: tables/ab_scan.tex
\begin{table}[t]
    \centering
    \renewcommand{\arraystretch}{1.1}
    \setlength{\tabcolsep}{1.6mm}
    \caption{Ablation study of local scan with LocalVim-T$^*$ (no scan direction search, $2\times2$ window size) on ImageNet.}
    \small
    \begin{tabular}{l|ccc|c}
    \toprule
    Model & Horizontal scan & Local scan & SCAttn & ACC\\
    \midrule
    Vim-T & \checkmark & & & 73.1\\
    Vim-T w/ local scan & & \checkmark & & 74.1\\
    LocalVim-T$^*$ w/o SCAttn & \checkmark & \checkmark & & 75.2\\
    \rowcolor{mygray}
    LocalVim-T$^*$ & \checkmark & \checkmark & \checkmark & 75.8\\
    \bottomrule
    \end{tabular}
    \label{tab:ab_scan}
\end{table}

%% file: tables/coco_vim.tex
\begin{table}[h]
    \centering
    \renewcommand{\arraystretch}{1.1}
    \setlength{\tabcolsep}{1.5mm}
    \caption{Object detection and instance segmentation results on COCO \textit{val} set. Vim \cite{zhu2024vision} did not report the parameters and FLOPs of the models.}
    \small
    \begin{tabular}{l|cc|ccc|ccc}
    \toprule
    Backbone & Params & FLOPs & AP$^\mathrm{b}$ & AP${}^\mathrm{b}_{50}$ & AP${}^\mathrm{b}_{75}$ & AP${}^\mathrm{b}_{s}$ & AP${}^\mathrm{b}_{m}$ & AP${}^\mathrm{b}_{l}$\\
    \midrule
    DeiT-Ti & - & - & 44.4 & 63.0 & 47.8 & 26.1 & 47.4 & 61.8\\
    Vim-Ti & - & - & 45.7 & 63.9 & 49.6 & 26.1 & 49.0 & 63.2 \\
    \rowcolor{mygray}
    LocalVim-T & 31M & 403G & 45.3 & 66.2 & 49.1 & 26.0 & 49.5 & 61.7 \\
    \midrule
    Backbone & Params & FLOPs & AP$^\mathrm{m}$ & AP${}^\mathrm{m}_{50}$ & AP${}^\mathrm{m}_{75}$ & AP${}^\mathrm{m}_s$ & AP${}^\mathrm{m}_{m}$ & AP${}^\mathrm{m}_{l}$\\
    \midrule
    DeiT-Ti & - & - & 38.1 & 59.9 & 40.5 & 18.1 & 40.5 & 58.4\\
    Vim-Ti & - & - & 39.2 & 60.9 & 41.7 & 18.2 & 41.8 & 60.2\\
    \rowcolor{mygray}
    LocalVim-T & 31M & 403G & 39.9 & 63.0 & 42.5 & 17.7 & 43.0 & 60.5\\
    \bottomrule
    \end{tabular}
    \label{tab:coco_vim}
\end{table}